\documentclass[11pt,a4paper]{article}
\usepackage[hyperref]{eacl2021}
\usepackage{times}
\usepackage{latexsym}

\usepackage{microtype}

\aclfinalcopy 

\title{Syntactic Nuclei in Dependency Parsing -- A Multilingual Exploration}

\author{Ali Basirat \\
  Uppsala University \\
  Dept.\ of Linguistics and Philology \\
  \texttt{ali.basirat@lingfil.uu.se} \\\And
  Joakim Nivre \\
  Uppsala University \\
  Dept.\ of Linguistics and Philology \\
  \texttt{joakim.nivre@lingfil.uu.se} \\}

\date{}

\usepackage{latexsym}

\usepackage[utf8]{inputenc}
\usepackage{amssymb}
\usepackage{amsmath}
\usepackage{natbib}
\setcitestyle{authoryear, open={(},close={)}}

\usepackage{tikz}
\usepackage{tikzscale}
\usepackage{pgfplots}
\usepgfplotslibrary{colorbrewer}
\usepgfplotslibrary{groupplots}
\usepackage{pgfplotstable}

\usepackage{graphicx}
\usepackage{caption}
\usepackage{subcaption}

\usepackage{multirow}
\usepackage{array}

\usepackage{comment}
\usepackage{enumitem}

\usepackage{CJKutf8}

\usepackage[]{xcolor}
\usepackage[underline=true,rounded corners=false]{pgf-umlsd}
\definecolor{cor-very-weak}{HTML}{BBBBBB}
\definecolor{cor-weak}{HTML}{EEBD84}
\definecolor{cor-moderate}{HTML}{F47461}
\definecolor{cor-strong}{HTML}{CB2F44}
\definecolor{cor-very-strong}{HTML}{8B0000}

\usepackage{rotating}

\usepackage{tikz-dependency}
\usepackage{fontenc}
\usepackage{lingmacros}

\begin{document}

\maketitle

\begin{abstract}
Standard models for syntactic dependency parsing take words to be the elementary units that enter into dependency relations. In this paper, we investigate whether there are any benefits from enriching these models with the more abstract notion of nucleus proposed by Tesni\`{e}re. We do this by showing how the concept of nucleus can be defined in the framework of Universal Dependencies and how we can use composition functions to make a transition-based dependency parser aware of this concept. Experiments on 12 languages show that nucleus composition gives small but significant improvements in parsing accuracy. Further analysis reveals that the improvement mainly concerns a small number of dependency relations, including nominal modifiers, relations of coordination, main predicates, and direct objects. 
\end{abstract}

\section{Introduction}

\begin{figure*}
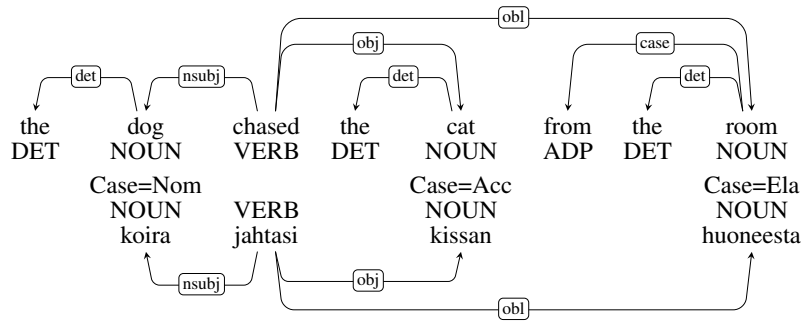

\centering
\scalebox{0.85}{
\begin{dependency}
\begin{deptext}[column sep=0.3cm,
]
the \& dog \& chased \& the \& cat \& from \& the \& room \\
\textcolor{black}{DET} \& \textcolor{black}{NOUN} \& \textcolor{black}{VERB} \& \textcolor{black}{DET} \& \textcolor{black}{NOUN} \& \textcolor{black}{ADP} \& \textcolor{black}{DET} \& \textcolor{black}{NOUN} \\[0.15cm]
\& \textcolor{black}{Case=Nom} \& \& \& \textcolor{black}{Case=Acc} \& \& \& \textcolor{black}{Case=Ela} \\
\& \textcolor{black}{NOUN} \& \textcolor{black}{VERB} \& \& \textcolor{black}{NOUN} \& \& \& \textcolor{black}{NOUN} \\
\& koira \& jahtasi \& \& kissan \& \& \& huoneesta \\
\end{deptext}
\depedge{2}{1}{det}
\depedge{3}{2}{nsubj}
\depedge[edge below]{3}{2}{nsubj}

\depedge{5}{4}{det}
\depedge{3}{5}{obj}
\depedge[edge unit distance=0.6em,edge below]{3}{5}{obj}

\depedge{8}{7}{det}
\depedge{8}{6}{case}
\depedge[edge unit distance=0.75em]{3}{8}{obl}
\depedge[edge unit distance=0.47em,edge below]{3}{8}{obl}
\end{dependency}
}
\caption{Word-based dependency trees for equivalent sentences from English (top) and Finnish (bottom).}
\label{fig:en-fi}
\end{figure*}

A syntactic dependency tree consists of directed arcs, representing syntactic relations like subject and object, connecting a set of nodes, representing the elementary syntactic units of a sentence. In contemporary dependency parsing, it is generally assumed that the elementary units are word forms or tokens, produced by a tokenizer or word segmenter. A consequence of this assumption is that the shape and size of dependency trees will vary systematically across languages. In particular, morphologically rich languages will typically have fewer elementary units and fewer relations than more analytical languages, which use independent function words instead of morphological inflection to encode grammatical information. This is illustrated in Figure~\ref{fig:en-fi}, which contrasts two equivalent sentences in English and Finnish, annotated with dependency trees following the guidelines of Universal Dependencies (UD) \citep{nivre16lrec,nivre20lrec}, which assume words as elementary units.

\begin{figure}[t]
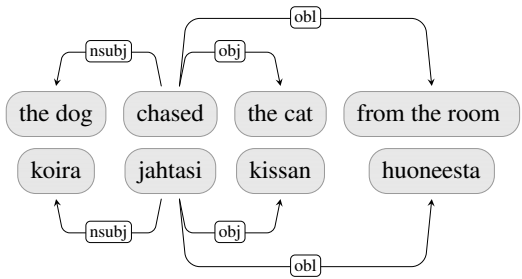

\centering
\scalebox{0.85}{
\depstyle{bubble}{draw=gray!80, minimum height=18pt, rounded corners=8pt,
   inner sep=3pt, top color=gray!20, bottom color=gray!20}
\begin{dependency}
\begin{deptext}[column sep=0.5cm,
]
the dog \& chased \& the cat \& from the room~ \\[0.5cm]
koira \& jahtasi \& kissan \& huoneesta \\
\end{deptext}
\wordgroup[bubble]{1}{2}{2}{en-pred}
\wordgroup[bubble]{2}{2}{2}{fi-pred}

\wordgroup[bubble]{1}{1}{1}{en-subj}
\groupedge{en-pred}{en-subj}{nsubj}{15}
\wordgroup[bubble]{2}{1}{1}{fi-subj}
\groupedge[edge below]{fi-pred}{fi-subj}{nsubj}{15}

\wordgroup[bubble]{1}{3}{3}{en-obj}
\groupedge{en-pred}{en-obj}{obj}{15}
\wordgroup[bubble]{2}{3}{3}{fi-obj}
\groupedge[edge below]{fi-pred}{fi-obj}{obj}{15}

\wordgroup[bubble]{1}{4}{4}{en-obl}
\groupedge[edge unit distance=0.4]{en-pred}{en-obl}{obl}{30}
\wordgroup[bubble]{2}{4}{4}{fi-obl}
\groupedge[edge below, edge unit distance=0.4]{fi-pred}{fi-obl}{obl}{30}
\end{dependency}
}
\caption{Nucleus-based dependency trees for equivalent sentences from English (top) and Finnish (bottom).}
\label{fig:en-fi-bubble}
\end{figure}

An alternative view, found in the seminal work of \citet{Lucien_Tesniere_1959}, is that dependency relations hold between slightly more complex units called \emph{nuclei}, semantically independent units consisting of a content word together with its grammatical markers, regardless of whether the latter are realized as independent words or not. Thus, a nucleus will often correspond to a single word -- as in the English verb \emph{chased}, where tense is realized solely through morphological inflection -- but it may also correspond to several words -- as in the English verb group \emph{has chased}, where tense is realized by morphological inflection in combination with an auxiliary verb. The latter type is known as a \emph{dissociated nucleus}. If we assume that the elementary syntactic units of a dependency tree are nuclei rather than word forms, then the English and Finnish sentences will have the same dependency trees, visualized in Figure~\ref{fig:en-fi-bubble}, and will differ only in the realization of their nuclei. In particular, while nominal nuclei in Finnish are consistently realized as single nouns inflected for case, the nominal nuclei in English involve standalone articles and the preposition \emph{from}.

In this paper, we set out to investigate whether research on dependency parsing can benefit from making explicit use of Tesni\`{e}re's notion of nucleus, from the point of view of accuracy, interpretability and evaluation. We do this from a multilingual perspective, because it is likely that the effects of introducing nuclei will be  different in different languages, and we strongly believe that a comparison between different languages is necessary in order to assess the potential usefulness of this notion. We are certainly not the first to propose that Tesni\`{e}re's notion of nucleus can be useful in parsing. One of the earliest formalizations of dependency grammar for the purpose of statistical parsing, that of \citet{samuelsson-2000-statistical}, had this notion at its core, and \citet{Sangati2009} presented a conversion of the Penn Treebank of English to Tesni\`{e}re style representations, including nuclei. However, previous attempts have been hampered by the lack of available parsers and resources to test this hypothesis on a large scale. Thus, the model of \citet{samuelsson-2000-statistical} was never implemented, and the treebank conversion of \citet{Sangati2009} is available only for English and in a format that no existing dependency parser can handle. We propose to overcome these obstacles in two ways. On the resource side, we will rely on UD treebanks and exploit the fact that, although the annotation is word-based, the guidelines prioritize dependency relations between content words that are the cores of syntactic nuclei, which facilitates the recognition of dissociated nuclei and gives us access to annotated resources for a wide range of languages. On the parsing side, we will follow a transition-based approach, which can relatively easily be extended to include operations that create representations of syntactic nuclei, as previously shown by \citet{de-lhoneux-etal-2019-recursive}, something that is much harder to achieve in a graph-based approach.

\section{Related Work}
Dependency-based guidelines for syntactic annotation generally discard the nucleus as the basic syntactic unit in favor of the (orthographic) word form, possibly with a few exceptions for fixed multiword expressions. A notable exception is the three-layered annotation scheme of the Prague Dependency Treebank \citep{HajicHajicovaAl2000}, where nucleus-like concepts are captured at the tectogrammatical level according to the Functional Generative Description \citep{sgall86}. 
\citet{barzdins-etal-2007-dependency} propose a syntactic analysis model for Latvian based on the x-word concept analogous to the nucleus concept. In this grammar, an x-word acts as a non-terminal symbol in a phrase structure grammar and can appear as a head or dependent in a dependency tree. \citet{Nespore2010ComparisonOT} compare this model to the original dependency formalism of \citet{Lucien_Tesniere_1959}. Finally, as already mentioned, \citet{Sangati2009} develop an algorithm to convert English phrase structure trees to 
Tesni\`{e}re style representations.

When it comes to syntactic parsing, \citet{jarvinen-tapanainen-1998-towards} were pioneers in adapting Tesni\`{e}re's dependency grammar for computational processing. They argue that the nucleus concept is crucial to establish cross-linguistically valid criteria for headedness and that it is not only a syntactic primitive but also the smallest semantic unit in a lexicographical description. As an alternative to the rule-based approach of \citet{jarvinen-tapanainen-1998-towards}, \citet{samuelsson-2000-statistical} defined a generative statistical model for nucleus-based dependency parsing, which however 
was never implemented.

The nucleus concept has affinities with the chunk concept found in many approaches to parsing, starting with \citet{abney91parsing}, who proposed to first find chunks and then dependencies between chunks, an idea that was generalized into cascaded parsing by \citet{buchholz99} among others. It is also clearly related to the vibhakti level in the Paninian computation grammar framework \citep{bharati93,bharati-etal-2009-two}.
In a similar vein, \citet{Kudo_chuncking_depparsing} use cascaded chunking for dependency parsing of Japanese,
\citet{tongchim-etal-2008-experiments} show that base-NP chunking can significantly improve the accuracy of dependency parsing for Thai, and \citet{durgar-el-kahlout-etal-2014-initial} show that chunking improves dependency parsing of Turkish. \citet{das-etal-2016-cross} study the importance of chunking in the transfer parsing model between Hindi and Bengali, and \citet{lacroix-2018-investigating} show that NP chunks are informative for universal part-of-speech tagging and dependency parsing. 

In a more recent study, \citet{delhoneux19nucleus} investigate whether the hidden representations of a neural transition-based dependency parser encodes information about syntactic nuclei, with special reference to verb groups. They find some evidence that this is the case, especially if the parser is equipped with a mechanism for recursive subtree composition of the type first proposed by \citet{stenetorp13} and later developed by \citet{dyer-etal-2015-transition} and \citet{de-lhoneux-etal-2019-recursive}.
The idea is to use a composition operator that recursively combines information from subtrees connected by a dependency relation into a representation of the new larger subtree. In this paper, we will exploit variations of this technique to create parser-internal representations of syntactic nuclei, as discussed in Section~\ref{sec:nucleus_composition}. However, first we need to discuss how to identify nuclei in UD treebanks.

\section{Syntactic Nuclei in UD}
\label{sec:treebanks}

UD\footnote{https://universaldependencies.org} \citep{nivre16lrec,nivre20lrec} is an ongoing project aiming to provide cross-linguistically consistent morphosyntactic annotation of many languages around the world. The latest release (v2.7) contains 183 treebanks, representing 104 languages and 20 language families. The syntactic annotation in UD is based on dependencies and the elementary syntactic units are assumed to be words, but the style of the annotation makes it relatively straightforward to identify substructures corresponding to (dissociated) nuclei. More precisely, UD prioritizes direct dependency relations between content words, as opposed to relations being mediated by function words, which has two consequences. First, incoming dependencies always go to the lexical core of a nucleus.\footnote{Except in some cases of ellipsis, like \emph{she did}, where the auxiliary verb \emph{did} is ``promoted'' to form a nucleus on its own.} Second, function words are normally leaves of the dependency tree, attached to the lexical core with special dependency relations, which we refer to as functional relations.\footnote{Again, there are a few well-defined exceptions to the rule that function words are leaves, including ellipsis, coordination, and fixed multiword expressions.}

\begin{figure*}[t]
    \centering
    \resizebox{\textwidth}{!}{%
        \includegraphics[]{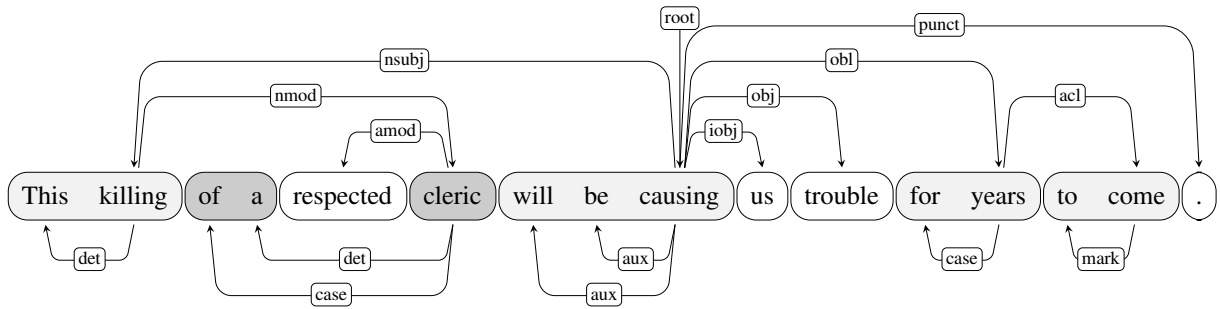}
    }
    \caption{Syntactic UD representation with functional relations drawn below the sentence. Dissociated nuclei are grayed, with a darker shade for the discontiguous nucleus.}
    \label{fig:nuclei_in_ud}
\end{figure*}

Figure~\ref{fig:nuclei_in_ud} illustrates these properties of UD representations by showing the dependency tree for the English sentence \emph{This killing of a respected cleric will be causing us trouble for years to come} with functional relations drawn below the sentence and other relations above. Given this type of representation, we can define a \emph{nucleus} as a subtree where all internal dependencies are functional relations, as indicated by the ovals in Figure~\ref{fig:nuclei_in_ud}. The nuclei can be divided into single-word nuclei, whitened, and dissociated nuclei, grayed. The latter can be contiguous or discontiguous, as shown by the nucleus \emph{of a cleric}, which consists of the two parts colored with a darker shade. 

This definition of nucleus in turn depends on what we define to be functional relations. For this study, we assume that the following 7 UD relations\footnote{A more detailed description of the relations is available in the UD documentation at https://universaldependencies.org.} belong to this class:
\begin{itemize}[itemsep=0pt,topsep=3pt]
    \item Determiner (\emph{det}): the relation between a determiner, mostly an article or demonstrative, and a noun. %
    Especially for articles, there is considerable cross-linguistic variation. 
    For example, definiteness is expressed by an independent function word in English (\emph{the girl}), by a morphological inflection in Swedish (\emph{flicka-n}), and not at all in Finnish. 
    \item Case marker (\emph{case}): the relation between a noun and a case marker when it is a separate syntactic word and not an affix. %
    UD takes a radical approach to adpositions and treats them all as case markers. Thus, in Figure~\ref{fig:en-fi}, we see that the English adposition \emph{from} corresponds to the Finnish elative case inflection. 
    \item Classifier (\emph{clf}): the relation between a classifier, a counting unit used for conceptual classification of nouns, and a noun. %
    This relation is seen in languages which have a classification system such as Chinese. %
    For example, English \emph{three students} corresponds to Chinese 
    \begin{CJK*}{UTF8}{gbsn}
    三个学生,
    \end{CJK*}
    literally ``three [human-classifier] student''.
    \item Auxiliary (\emph{aux}): the relation between an auxiliary verb or nonverbal TAME marker and a verbal predicate. %
    An example is the English verb group \emph{will be causing} in Figure~\ref{fig:nuclei_in_ud}, which alternates with finite main verbs like \emph{causes} and \emph{caused}. 
    \item Copula (\emph{cop}): the relation between a verbal or nonverbal copula and a nonverbal predicate. %
    For example, in English \emph{Ivan is the best dancer}, the copula \emph{is} links the predicate \emph{the best dancer} to \emph{Ivan}, but it has no counterpart in Russian \emph{Ivan luc\v{s}\v{\i}j tancor}, literally ``Ivan best dancer''. 
    \item Subordination marker (\emph{mark}): the relation between a subordinator and the predicate of a subordinate clause. %
    This is exemplified by the infinitive marker \emph{to} in Figure~\ref{fig:nuclei_in_ud}. Other examples are subordinating conjunctions like \emph{if}, \emph{because} and \emph{that}, the function of which may be encoded morphologically or through word order in other languages.
    \item Coordinating conjunction (\emph{cc}): the relation between a coordinator and a conjunct (typically the last one) in a coordination. %
    Thus, in \emph{apples, bananas and oranges}, UD treats \emph{and} as a dependent of \emph{oranges}. This linking function may be missing or expressed morphologically in other languages.
\end{itemize} 
The inclusion of the \emph{cc} relation among the nucleus-internal relations is probably the most controversial decision, given that Tesni\`{e}re treated coordination (including coordinating conjunctions) as a third type of grammatical relation -- junction (fr.\ jonction) -- distinct from both dependency relations and nucleus-internal relations. However, 
we think coordinating conjunctions have enough in common with other function words to be included in this preliminary study and leave further division into finer categories for future work.\footnote{In addition to separating the \emph{cc} relation from the rest, such a division might include distinguishing nominal nucleus relations (\emph{det}, \emph{case} and \emph{clf}) from predicate nucleus relations (\emph{aux}, \emph{cop} and \emph{mark}).}

Given the definition of nucleus in terms of functional UD relations, it would be straightforward to convert the UD representations to dependency trees where the elementary syntactic units are nuclei rather than words. However, the usefulness of such a resource would currently be limited, given that it would require parsers that can deal with nucleus recognition, either in a preprocessing step or integrated with the construction of dependency trees, and such parsers are not (yet) available. Moreover, evaluation results would not be comparable to previous research. Therefore, we will make use of the nucleus concept in UD in three more indirect ways:
\begin{itemize}[itemsep=0pt,topsep=3pt]
\item \textbf{Evaluation:} Even if a parser outputs a word-based dependency tree in UD format, we can evaluate its accuracy on nucleus-based parsing by simply not scoring the functional relations. This is equivalent to the Content Labeled Attachment Score (CLAS) previously proposed by \citet{nivre17udw}, and we will use this score as a complement to the standard Labeled Attachment Score (LAS) in our experiments.\footnote{Our use of CLAS differs only in that we include punctuation in the evaluation, whereas \citet{nivre17udw} excluded it.}
  
\item \textbf{Nucleus Composition:} Given our definition of nucleus-internal relations, we can make parsers aware of the nucleus concept by differentiating the way they predict and represent dissociated nuclei and dependency structures, respectively. More precisely, we will make use of composition operations to create internal representations of (dissociated) nuclei, as discussed in detail in Section~\ref{sec:nucleus_composition} below.

\item \textbf{Oracle Parsing:} To establish an upper bound on what a nucleus-aware parser can achieve, we will create a version of the UD representation which is still a word-based dependency tree, but where nuclei are explicitly represented by letting the word form for each nucleus core be a concatenation of all the word forms that are part of the nucleus.\footnote{The English sentence in Figure~\ref{fig:en-fi} thus becomes: \emph{the dog-the chased the cat-the from the room-the-from}.} 
We call this oracle parsing to emphasize that the parser has oracle information about the nuclei of a sentence, although it still has to predict all the syntactic relations. 
\end{itemize}

\section{Syntactic Nuclei in Transition-Based Dependency Parsing}
\label{sec:nucleus_composition}
A transition-based dependency parser derives a dependency tree from the sequence of words forming a sentence \citep{yamada-matsumoto-2003-statistical,nivre-2003-efficient,nivre-2004-incrementality}. The parser constructs the tree incrementally by applying transitions, or parsing actions, to configurations consisting of a stack $S$ of partially processed words, a buffer $B$ of remaining input words, and a set of dependency arcs $A$ representing the partially constructed dependency tree. 
The process of parsing starts from an initial configuration and ends when the parser reaches a terminal configuration.
The transitions between configurations are predicted by a history-based model that combines information from 
$S$, $B$ and $A$.

For the experiments in this paper, we use a version of the arc-hybrid transition system initially proposed by \citet{kuhlmann-etal-2011-dynamic}, where the initial configuration has all words $w_1, \ldots, w_n$ plus an artificial root node $r$ in $B$, while $S$ and $A$ are empty.\footnote{Positioning the artificial root node at the end of the buffer is a modification of the original system by \citet{kiperwasser-goldberg-2016-simple}, inspired by the results reported in \citet{BallesterosGTT13}.} 
There are four transitions: Shift, Left-Arc, Right-Arc and Swap. 
Shift pushes the first word $b_0$ in $B$ onto $S$ (and is not permissible if $b_0 = r$). Left-Arc attaches the top word $s_0$ in $S$ to $b_0$ and removes $s_0$ from $S$, while Right-Arc attaches $s_0$ to the next word $s_1$ in $S$ and removes $s_0$ from $S$. Swap, finally, moves $s_1$ back to 
$B$ in order to allow the construction of non-projective dependencies.\footnote{This extension of the arc-hybrid system was proposed by  \citet{de-lhoneux-etal-2017-arc}, inspired by the corresponding extension of the arc-standard system by \citet{nivre09acl}.}

Our implementation of this transition-based parsing model is based on the influential architecture of \citet{kiperwasser-goldberg-2016-simple}, which takes as input a sequence of vectors $x_1, \ldots, x_n$ representing the input words $w_1, \ldots, w_n$ and feeds these vectors through a BiLSTM that outputs contextualized word vectors $v_1, \ldots, v_n$, which are stored in the buffer $B$. Parsing is then performed by iteratively applying the transition predicted by an MLP taking as input a small number of contextualized word vectors from the stack $S$ and the buffer $B$. More precisely, in the experiments reported in this paper, the predictions are based on the two top items $s_0$ and $s_1$ in $S$ and the first item $b_0$ in $B$. In a historical perspective, this may seem like an overly simplistic prediction model, but recent work has shown that more complex feature vectors are largely superfluous thanks to the BiLSTM encoder \citep{shi17emnlp,falenska19}.

The transition-based parser as described so far does not provide any mechanism for modeling the nucleus concept. It is a purely word-based model, where any more complex syntactic structure is represented internally by the contextualized vector of its head word. Specifically, when two substructures $h$ and $d$ are combined in a Left-Arc or Right-Arc transition, only the vector $v_h$ representing the syntactic head is retained in $S$ or $B$, while the vector $v_d$ representing the syntactic dependent is removed from $S$. In order to make the parser sensitive to (dissociated) nuclei in its internal representations, we follow \citet{de-lhoneux-etal-2019-recursive} and augment the Right-Arc and Left-Arc actions with a composition operation. The idea is that, whenever the substructures $h$ and $d$ are combined with 
label $l$, we replace the current representation of $h$ with the output of a function $f(h, d, l)$. We can then control the information flow for nuclei and other 
constructions through the definition of $f(h, d, l)$.

\paragraph{Hard Composition:}
The simplest version, which we call \emph{hard} composition, is to explicitly condition the composition on the dependency label $l$. In this setup, $f(h, d, l)$ combines the head and dependent vectors only if $l$ is a functional relation and simply returns the head vector otherwise:
\begin{equation}
    f(h, d, l) = 
    \begin{cases}
        \Vec{h} \circ \Vec{d} & \text{if } l\in F \\
        \Vec{h} & \text{otherwise}
    \end{cases}
    \label{eq:hard_composition}
\end{equation}
We use $\Vec{x}$ to denote the vector representation of $x$\footnote{In the baseline parser, $\Vec{h}$ is always identical to the contextualized representation $v_h$ of the head word $w_h$, but after introducing composition operations we need a more abstract notation.} and $F$ to denote the set of seven functional relations defined in Section~\ref{sec:treebanks}. The composition operator $\circ$ can be any function of the form $\mathbb{R}^n\times\mathbb{R}^n\to\mathbb{R}^n$, e.g., vector addition $\Vec{h}+\Vec{d}$, where $n$ is the dimensionality of the vector space. 

\paragraph{Soft Composition:} 
The \emph{soft} composition is similar to the hard composition, but instead of applying the composition operator to the head and dependent vectors, the operator is applied to the head vector and a vector representation of the entire dependency arc $(h, d, l)$. The vector representation of the dependency arc is trained by a differentiable function $g$ that encodes the dependency label $l$ into a vector $\Vec{l}$ and maps the triple $(\Vec{h}, \Vec{d}, \Vec{l})$ to a vector space, i.e., $g: \mathbb{R}^n\times\mathbb{R}^n\times\mathbb{R}^m\to\mathbb{R}^n$ where $n$ and $m$ are the dimensionalities of the word and label spaces, respectively. An example of 
$g$ is a perceptron with a sigmoid activation that maps the vector representations of $h$, $d$ and $l$ to a vector space:  
\begin{equation}
    g(h, d, l) = \sigma(W(\Vec{h}\odot\Vec{d}\odot\Vec{l})+b)
    \label{eq:soft_composition}
\end{equation}
where $\odot$ is the vector concatenation operator. The soft nucleus composition is then:
\begin{equation}
     f(h, d, l) = 
    \begin{cases}
        \Vec{h}\circ g(h,d,l), & \text{if } l\in F \\
        \Vec{h} & \text{otherwise}
    \end{cases}
    \label{eq:soft_composition_functional}   
\end{equation}
The parameters of the function $g$ are trained with the other parameters of the parser. 

\paragraph{Generalized Composition:} To test our hypothesis that composition is beneficial for dissociated nuclei, we contrast both hard and soft composition to a generalized version of soft composition, where we do not restrict the application to functional relations. In this case, the composition function 
is:
\begin{equation}
    f(h,d,l)=\Vec{h}\circ g(h,d,l)
    \label{eq:mechanized_soft_compositition}
\end{equation}
where $l$ can be any dependency label. 
In this approach, the if-clause in Equation~\ref{eq:soft_composition} and \ref{eq:soft_composition_functional} is eliminated and the parser itself learns in what conditions the composition should be performed. In particular, if the composition operator is addition, and $g$ is a perceptron with a sigmoid activation on the output layer (as in Equation~\ref{eq:soft_composition}), then $g$ operates as a gate that controls the contribution of the dependency elements $h$, $d$, and $l$ to the composition. If the composition should not be performed, it returns a vector close to zero. 

\input{data.tex}
\section{Experiments}
In the previous sections, we have shown how syntactic nuclei can be identified in the UD annotation and how transition-based parsers can be made sensitive to these structures in their internal representations through the use of nucleus composition. We now proceed to a set of experiments investigating the impact of nucleus composition on a diverse selection of languages. 

\begin{table*}[t]
    \centering
    \renewcommand{\tabcolsep}{2.5pt}
    \begin{tabular}{l|l|l|r||r|r|r|r|r||r|r|r|r|r}
         \multicolumn{4}{c||}{} & \multicolumn{5}{c||}{ LAS } & \multicolumn{5}{c}{ CLAS } \\\hline
        Language & Treebank & Family & Size & Base & Hard & Soft & Gen & Ora & Base & Hard & Soft & Gen & Ora \\\hline
        Arabic & PADT & AA-Semitic & $242$K & 78.0 & 78.1 & 78.3 & 78.2 & 80.5 & 74.4 & 74.6 & 74.9 & 74.8 & 75.8 \\
        Basque & BDT & Basque & $70$K & 73.5 & 73.4 & 73.9 & 74.2 & 78.9 & 70.9 & 70.9 & 71.3 & 71.8 & 75.3 \\
        Chinese & GSD & Sino-Tibetan & $49$K & 72.1 & 72.1 & 72.7 & 72.2 & 80.4 & 68.8 & 68.7 & 69.4 & 69.0 & 74.6 \\
        English & EWT & IE-Germanic & $188$K & 82.5 & 82.4 & 82.5 & 82.4 & 85.6 & 78.8 & 78.7 & 78.8 & 78.7 & 81.0 \\
        Finnish & TDT & Uralic & $168$K & 78.5 & 78.9 & 79.6 & 79.9 & 84.2 & 77.5 & 77.8 & 78.6 & 78.9 & 81.8 \\
        Hebrew & HTB & AA-Semitic & $108$K & 81.5 & 81.6 & 81.8 & 82.1 & 83.3 & 74.5 & 74.7 & 74.9 & 75.3 & 75.9 \\
        Hindi & HDTB & IE-Indic & $294$K & 87.9 & 88.1 & 88.4 & 89.0 & 89.8 & 83.9 & 83.9 & 84.4 & 85.4 & 85.0 \\
        Italian & ISDT & IE-Romance & $252$K & 87.4 & 87.6 & 87.9 & 87.5 & 89.0 & 81.5 & 81.6 & 82.1 & 81.6 & 82.8 \\
        Japanese & GSD & Japanese & $61$K & 93.4 & 93.4 & 93.5 & 93.4 & 94.1 & 89.7 & 89.7 & 89.9 & 89.7 & 90.5 \\
        Korean & GSD & Korean & $21$K & 75.1 & 75.0 & 75.4 & 75.6 & 76.6 & 74.9 & 74.8 & 75.2 & 75.4 & 75.7 \\
        Swedish & Talbanken & IE-Germanic & $61$K & 76.9 & 77.3 & 77.5 & 77.6 & 82.9 & 73.2 & 73.5 & 73.8 & 74.1 & 78.2 \\
        Turkish & IMST & Turkic & $40$K & 55.6 & 55.3 & 56.2 & 54.8 & 58.6 & 53.4 & 53.0 & 54.1 & 52.6 & 54.8 \\\hline
        Average & & & & 78.5 & 78.6 & 79.0 & 78.9 & 82.0 & 75.1 & 75.2 & 75.6 & 75.6 & 77.6 \\
    \end{tabular}
    \caption{Parsing accuracy for 5 parsing models evaluated on 12 UD treebanks. Language family includes genus according to WALS for large families (AA = Afro-Asian, IE = Indo-European). LAS = Labeled Attachment Score. CLAS = Content Labeled Attachment Score.}
    \label{tab:selected_languages}
\end{table*}

\subsection{Experimental Settings}
\label{sec:uuparser}
We use UUParser \citep{de-Lhoneux2017,smith-etal-2018-82}, an evolution of the transition-based dependency parser of \citet{kiperwasser-goldberg-2016-simple}, which was the highest ranked transition-based dependency parser in the CoNLL shared task on universal dependency parsing in 2018 \citep{zeman18conll}. 
As discussed in Section~\ref{sec:nucleus_composition}, this is a greedy transition-based parser based on the extended arc-hybrid system of \citet{de-lhoneux-etal-2017-arc}. It uses an MLP with one hidden layer to predict transitions between parser configurations, based on vectors representing two items on the stack $S$ and one item in the buffer $B$. In the baseline model, these items are contextualized word representations produced by a BiLSTM with two hidden layers. The input to the BiLSTM for each word is the concatenation of a randomly initialized word embedding and a character-based representation produced by running a BiLSTM over the character sequence of the word. We use a dimensionality of 100 for the word embedding as well as for the output of the character BiLSTM. 

For parsers with composition, we considered various composition operators $\circ$ and functions $g$. For the former, we tested vector addition, vector concatenation,\footnote{The concatenation operator requires special care to keep vector dimensionality constant. We double the dimensionality of the contextual vectors and fill the extra dimensions with zeros. We then replace the zero part of the second operand with the first operand's non-zero part at composition time.} and perceptron. For the latter we tried a multi-layer perceptron with different activation functions. Based on the results of the preliminary experiments, we selected vector addition for the composition operator $\circ$ and the perceptron with sigmoid activation for the soft composition function $g$.
The inputs to the perceptron consist of two token vectors of size $512$ and a relation vector of size $10$. The token vectors are the outputs of the BiLSTM layer of the parser and the relation vector is trained by a distinct embedding layer. 

All parsers are trained for $50$ epochs and all reported results are averaged over 10 runs with different random seeds. 
Altogether we explore five different parsers:
\begin{itemize}[itemsep=0pt,topsep=3pt]
    \item \textbf{Base(line):} No composition.
    \item \textbf{Hard:} Baseline 
    + hard composition.
    \item \textbf{Soft:} Baseline 
    + soft composition.
    \item \textbf{Gen(eralized):} Baseline 
    + gen. 
    composition.
    \item \textbf{Ora(cle):} Baseline 
    trained and tested on 
    explicit annotation of nuclei 
    (see Section~\ref{sec:treebanks}).
\end{itemize}

\noindent
Our experiments are carried out on a typologically diverse set of languages with different degrees of morphosyntactic complexity, as shown in Table~\ref{tab:selected_languages}. The corpus size is the total number of words in each treebank. 
We use UD v2.3 with standard data splits \citep{ud_v23}. All evaluation results are on the development sets.\footnote{Since we want to perform an informative error analysis, we avoid using the dedicated test sets.}

\begin{figure*}[t]
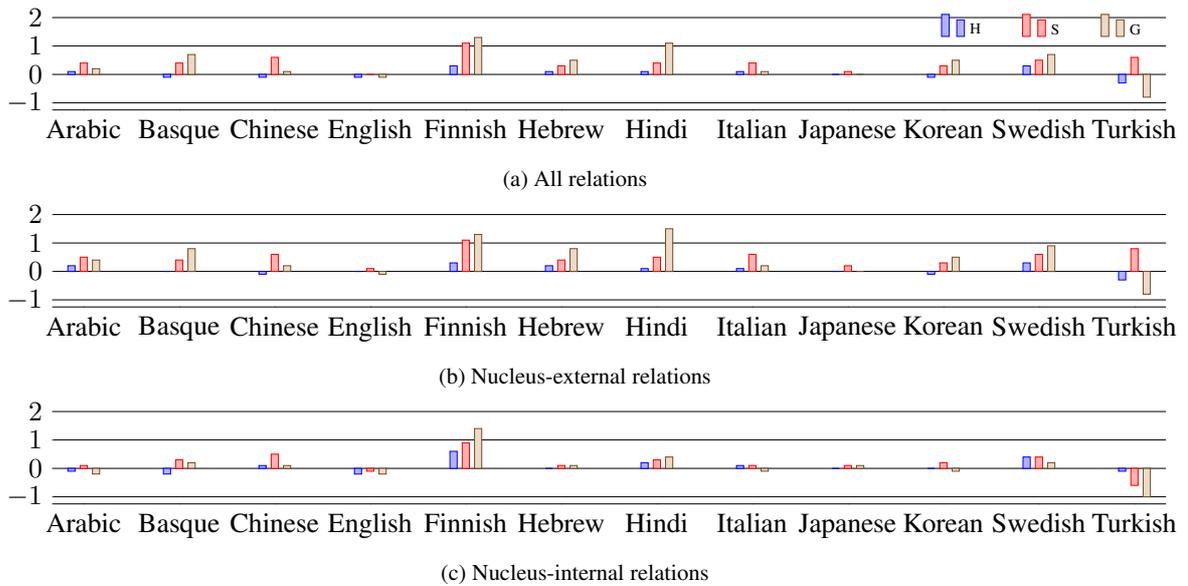

    \centering
\scalebox{0.95}{
    \begin{subfigure}{\textwidth}
        \includegraphics{results_composition_las.tikz}
        \caption{All relations}
        \label{fig:las_imp}
    \end{subfigure}
}\\
\scalebox{0.95}{
\begin{subfigure}{\textwidth}
        \includegraphics{results_composition_clas.tikz}
        \caption{Nucleus-external relations}
        \label{fig:clas_imp}
    \end{subfigure}
}\\
\scalebox{0.95}{    \begin{subfigure}{\textwidth}
        \includegraphics{results_composition_flas.tikz}
        \caption{Nucleus-internal relations}
        \label{fig:flas_imp}
    \end{subfigure}   
}
    \caption{Absolute improvement (or degradation) in labeled F-score with respect to the baseline for hard (H), soft (S) and generalized (G) composition for different sets of relations: (a) all relations (corresponding to LAS scores), (b) nucleus-external relations (corresponding to CLAS scores), (c) nucleus-internal relations.}
    \label{fig:summary}
\end{figure*}

\begin{figure*}[t]
    \centering
    
\scalebox{0.9}{        \includegraphics[width=\linewidth]{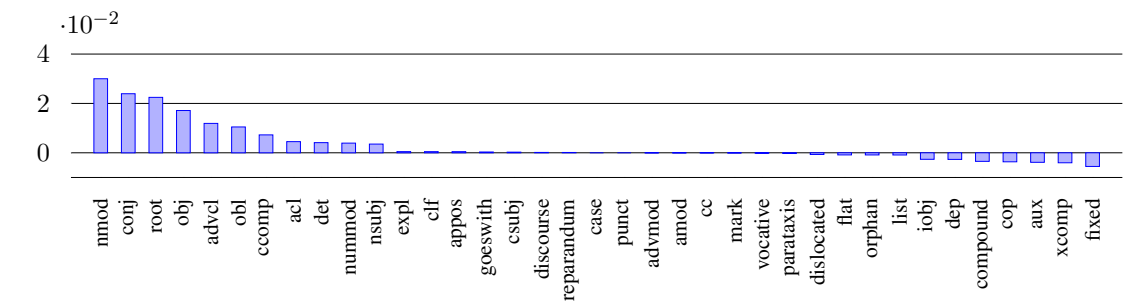}
}
    \caption{Improvement (or degradation) in labeled F-score with respect to the baseline for different UD relations (soft composition), weighted by the relative frequency of each relation and averaged across all languages.}
    \label{fig:score_imp}
\end{figure*}

\subsection{Results}
\label{sec:results}

Table~\ref{tab:selected_languages} reports the parsing accuracy achieved with our 5 parsers on the 12 different languages, using the standard LAS metric as well as the nucleus-aware CLAS metric. First of all, we see that hard composition is not very effective and mostly gives results in line with the baseline parser, except for small improvements for Finnish, Hindi and Swedish and a small degradation for Turkish. These differences are statistically significant for all four languages with respect to LAS but only for Finnish and Turkish with respect to CLAS (two-tailed $t$-test, $\alpha = .05$). By contrast, soft composition improves accuracy for all languages except English and the improvements are statistically significant for both LAS and CLAS. The average improvement is 0.5 percentage points for both LAS and CLAS, which indicates that most of the improvement occurs on nucleus-external relations thanks to a more effective internal representation of dissociated nuclei. There is some variation across languages, but the CLAS improvement is in the range 0.2--0.7 for most languages, with Finnish as the positive exception (1.1) and English as the negative one (0.0). Generalized composition, finally, where we allow composition also for non-functional relations, yields results very similar to those for soft composition, which could be an indication that the parser learns to apply composition mostly for functional relations. The results are a little less stable, however, with degradations for English and Turkish, and non-significant improvements for Chinese, Italian and Japanese. A tentative conclusion is therefore that composition is most effective when restricted to (but not enforced for) nucleus-internal relations.

Before we try to analyze the results in more detail, it is worth noting that most of the improvements due to composition are far below the improvements of the oracle parser.\footnote{The exception is generalized composition for Hindi, which exceeds the corresponding oracle parser with respect to CLAS.} However, it is important to keep in mind that, whereas the behavior of a composition parser is only affected \emph{after} a nucleus has been constructed, the oracle parser improves also with respect to the prediction of the nuclei themselves. This explains why the oracle parser generally improves more with respect to LAS than CLAS, and sometimes by a substantial margin (2.5 points for Chinese, 1.4 points for Basque and 1.3 points for Swedish).

\begin{table*}[t]
    \centering
    \begin{tabular}{lr|lr|lr|lr|lr}
        \multicolumn{2}{c}{ Finnish } & \multicolumn{2}{c}{ Chinese } & \multicolumn{2}{c}{ Swedish } & \multicolumn{2}{c}{ Turkish } & \multicolumn{2}{c}{ Hindi } \\\hline
        conj & 0.09 & det & 0.03 & xcomp & 0.07 & obj & 0.12 & obj & 0.03 \\
        root & 0.07 & case & 0.02 & advmod & 0.05 & conj & 0.10 & compound & 0.02 \\
        nmod & 0.06 & cop & 0.01 & nmod & 0.05 & root & 0.05 & nmod & 0.02 \\
        obl & 0.06 & conj & 0.01 & acl & 0.04 & acl & 0.04 & case & 0.02 \\
        ccomp & 0.03 & clf & 0.01 & conj & 0.04 & nummod & 0.03 & aux & 0.01 \\
        acl & 0.03 & ccomp & 0.00 & obl & 0.04 & obl & 0.03 & det & 0.00 \\
        obj & 0.02 & dep & 0.00 & obj & 0.04 & nmod & 0.01 & nummod & 0.00 \\
        xcomp & 0.01 & aux & 0.00 & mark & 0.03 & ccomp & 0.01 & amod & 0.00 \\
        nsubj & 0.01 & advcl & 0.00 & nsubj & 0.02 & det & 0.01 & advmod & 0.00 \\
        amod & 0.01 & cc & 0.00 & amod & 0.02 & cc & 0.01 & advcl & 0.00 \\
        \end{tabular}
    \caption{Improvement (or degradation) in labeled F-score, weighted by relative frequency, for the 10 best UD relations in the 5 languages with greatest LAS improvements over the baseline (soft composition).}
    \label{table:odds_values}
\end{table*}


Figure~\ref{fig:summary} visualizes the impact of hard, soft and generalized nucleus composition for different languages, with a breakdown into (a) all relations, which corresponds to the difference in LAS compared to the baseline, (b) nucleus-external relations, which corresponds to the difference in CLAS, and (c) nucleus-internal relations. Overall, these graphs are consistent with the hypothesis that using composition to create parser-internal representations of (dissociated) nuclei primarily affects the prediction of nucleus-external relations, as the (a) and (b) graphs are very similar and the (c) graphs mostly show very small differences. There are, however, two notable exceptions. For Finnish, all three composition methods clearly improve the prediction of nucleus-internal relations as well as nucleus-external relations, by over 1 F-score point for generalized composition. Conversely, for Turkish, especially the soft versions of composition has a detrimental effect on the prediction of nucleus-internal relations, reaching 1 F-score point for generalized composition. Turkish is also exceptional in showing opposite effects overall for soft and generalized composition, the former having a positive effect and the latter a negative one, whereas all other languages either show consistent trends or fluctuations around zero. Further research will be needed to explain what causes these deviant patterns.

Figure~\ref{fig:score_imp} shows the improvement (or degradation) for individual UD relations, weighted by relative frequency and averaged over all languages, for 
the best performing soft composition parser.
The most important improvements are observed
for \emph{nmod}, \emph{conj}, \emph{root} and \emph{obj}. The \emph{nmod} relation covers all nominal modifiers inside noun phrases, including prepositional phrase modifiers; the \emph{conj} relation holds between conjuncts in a coordination structure; the \emph{root} relation is assigned to the main predicate of a sentence; and \emph{obj} is the direct object relation. In addition, we see smaller improvements for a number of relations, including major clause relations like \emph{advcl} (adverbial clauses), \emph{obl} (oblique modifiers), \emph{ccomp} (complement clauses), and \emph{nsubj} (nominal subjects), as well as noun phrase internal relations like \emph{acl} (adnominal clauses, including relative clauses), \emph{det} (determiner), and \emph{nummod} (numeral modifier). Of these, only \emph{det} is a nucleus-internal relation, so the results further support the hypothesis that richer internal representations of (dissociated) nuclei primarily improve the prediction of nucleus-external dependency relations, especially major clause relations. 

It is important to remember that the results in Figure~\ref{fig:score_imp} are averaged over all languages and may hide interesting differences between languages. A full investigation of this variation is beyond the scope of this paper, but Table~\ref{table:odds_values} presents a further zoom-in by presenting statistics on the top 10 relations in the 5 languages where LAS improves the most compared to the baseline. To a large extent, we find the same relations as in the aggregated statistics, but there are also interesting language-specific patterns. For Chinese the top three relations (\emph{det}, \emph{case}, \emph{cop}) are all nucleus-internal relations; for Swedish the two top relations are \emph{xcomp} (open clausal complements) and \emph{advmod} (adverbial modifiers), neither of which show positive improvements on average; and for Hindi the \emph{compound} relation shows the second largest improvement. These differences definitely deserve further investigation.

\section{Conclusion}
We have explored how the concept of syntactic nucleus can be used to enrich the representations of a transition-based dependency parser, relying on UD treebanks for supervision and evaluation in experiments on a wide range of languages. We conclude that the use of composition operations for building internal representations of syntactic nuclei, in particular the technique that we have called soft composition, can lead to small but significant improvements in parsing accuracy for nucleus-external relations, notably for nominal modifiers, relations of coordination, main predicates, and direct objects. In future work we want to study the behavior of different types of nuclei in more detail, in particular how the different internal relations of nominal and verbal nuclei contribute to overall parsing accuracy. We also want to analyze the variation between different languages in more detail and see if it can be explained in terms of typological properties.

\section*{Acknowledgments}
We thank Daniel Dakota, Artur Kulmizev, Sara Stymne and Gongbo Tang for useful discussions and the EACL reviewers for constructive criticism. We acknowledge the computational resources provided by CSC in Helsinki and Sigma2 in Oslo through NeIC-NLPL (www.nlpl.eu).  

\bibliography{references}

\end{document}